\documentclass{bmvc2k}

\title{Hierarchical Contrastive Motion Learning for Video Action Recognition}

\addauthor{Xitong Yang}{xyang35@cs.umd.edu}{1}
\addauthor{Xiaodong Yang}{https://xiaodongyang.org}{2}
\addauthor{Sifei Liu}{https://www.sifeiliu.net}{2}
\addauthor{Deqing Sun}{https://deqings.github.io}{3}
\addauthor{Larry Davis}{http://users.umiacs.umd.edu/~lsd}{1}
\addauthor{Jan Kautz}{https://jankautz.com}{2}

\addinstitution{
 University of Maryland, College Park
}
\addinstitution{
 NVIDIA
}
\addinstitution{
 Google
}

\runninghead{Yang ET AL.}{Hierarchical Contrastive Motion Learning}

\usepackage{times}
\usepackage{epsfig}
\usepackage{graphicx}
\usepackage{amsmath}
\usepackage{amssymb}
\usepackage{bbm}

\usepackage{verbatim}
\usepackage{enumitem}
\usepackage{multirow}
\usepackage{xcolor}
\usepackage{colortbl}
\usepackage{booktabs}

\usepackage{floatrow}
\newfloatcommand{capbtabbox}{table}[][\FBwidth]

\usepackage{amssymb}
\usepackage{pifont}
\newcommand{\cmark}{\ding{51}}%
\newif\ifdrafting
\draftingtrue 
\ifdrafting
    \newcommand{\ds}[1]{{\leavevmode\color[rgb]{1,0,0}[Deqing: #1]}}
\else
    \newcommand{\ds}[1]{}
\fi

\begin{document}

\maketitle
\vspace{-0.07in}
\begin{abstract}
One central question for video action recognition is how to model motion. In this paper, we present hierarchical contrastive motion learning, a novel self-supervised learning framework to extract effective motion representations from raw video frames. Our approach progressively learns a hierarchy of motion features that correspond to different abstraction levels in a network. 
At each level, an explicit motion self-supervision is provided via contrastive learning to enforce the motion features to capture semantic dynamics and evolve  discriminatively for video action recognition.
This hierarchical design bridges the semantic gap between low-level movement cues and high-level recognition tasks, and promotes the fusion of appearance and motion information at multiple levels.
Our motion learning module is lightweight and flexible to be embedded into various backbone networks. Extensive experiments on four benchmarks show that our approach compares favorably against the state-of-the-art methods yet without requiring optical flow or supervised pre-training.
\end{abstract}

\section{Introduction}
Motion provides abundant and powerful cues for understanding the dynamic visual world.
A broad range of video understanding tasks 
benefit from the introduction of motion information, such as action recognition~\cite{corrnet,simonyan2014two}, activity detection~\cite{budget-aware,step}, object tracking~\cite{tracking20cvpr,simtrack}, etc.
Thus, how to extract and model motion is one of the fundamental problems in video understanding.
While early methods in this field mostly rely on the pre-computed motion features such as optical flow, recent works have been actively exploiting convolutional neural networks (CNNs) for more effective motion learning from raw video frames~\cite{ng2018actionflownet,corrnet}, encouraged by the success of end-to-end learning in various vision tasks.

A key challenge for end-to-end motion representation learning is to design an effective supervision. Unlike many other tasks that afford plenty of well-defined annotations, the ``ground truth motion'' is often unavailable or even undefined for motion learning in practice.
One popular idea is to extract motion features by means of the action recognition supervision.
However, the classification loss is shown to be sub-optimal in this task as it only provides an implied supervision to guide motion learning~\cite{stroud2018d3d}. 
This supervision is also prone to be biased towards appearance information since some video action benchmarks can be mostly solved by considering static images without temporal modeling~\cite{sevilla2019only}.
Recently, some efforts have been made to explore pretext tasks with direct supervisions for motion learning, such 
as optical flow prediction~\cite{ng2018actionflownet} and video frame reconstruction~\cite{zhu2018hidden}. 
Although having shown promising results, such supervisions are restricted to pixel-wise and short-term motion as they hinge on pixel photometric loss and movement between adjacent frames.

\begin{figure}[t]
\centering
\includegraphics[width=\linewidth]{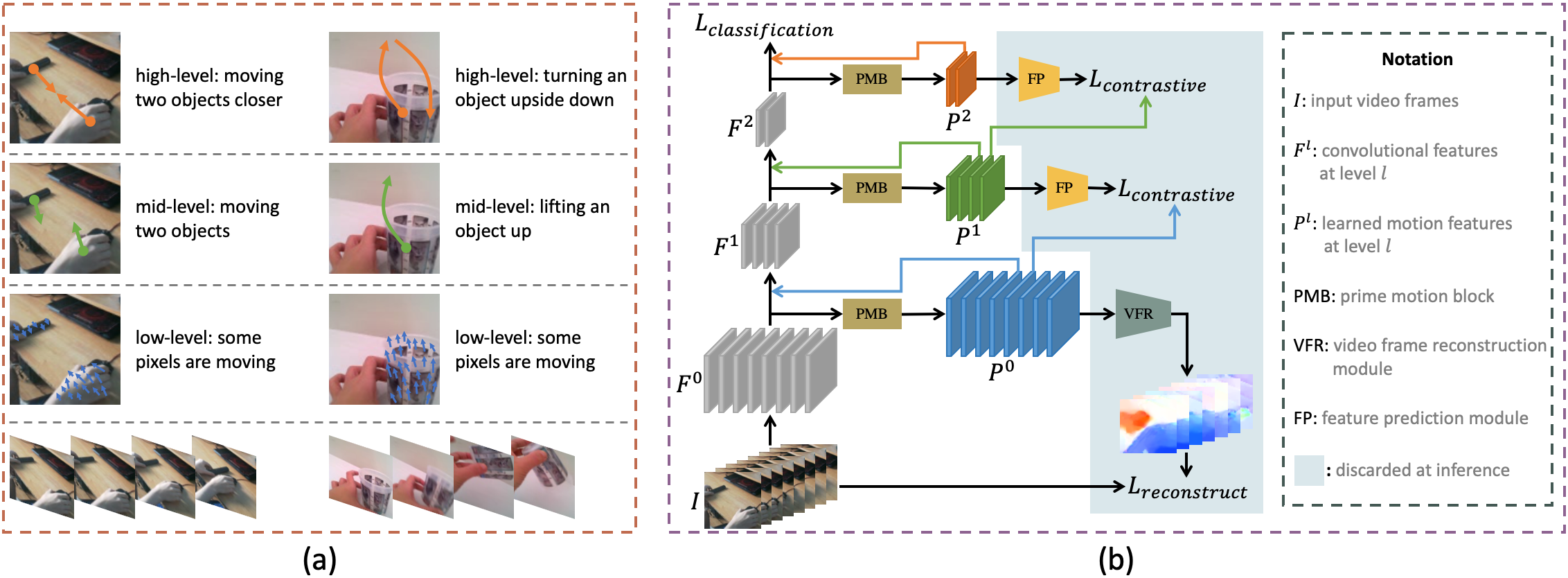}
\caption{(a) Illustration of the hierarchy of progressively learned motion: from pixel-level and short-range movement to semantic temporal dynamics. (b) Overview of the architecture of the proposed hierarchical contrastive motion learning framework.\vspace{-2mm}} 
\label{fig:teaser}\vspace{-2mm}
\end{figure}

In light of the above observations, we introduce a novel self-supervised learning framework that enables explicit motion supervision at multiple feature abstraction levels, which we term hierarchical contrastive motion learning. 
Specifically, given preliminary motion cues as a bootstrap, our approach progressively learns a hierarchy of motion features in a bottom-up manner, as illustrated in Figure~\ref{fig:teaser}(a). 
This hierarchical design is proposed to bridge the semantic gap between the low-level preliminary motion and the high-level recognition task---analogous to the findings in neuroscience that humans perceive motion patterns in a hierarchical way~\cite{giese2003neural,bill2019hierarchical}.
At each level, a discriminative contrastive loss~\cite{chopra2005learning,hadsell2006dimensionality} 
provides an explicit self-supervision to enforce the motion features at current level to predict the future ones at previous level.
In contrast to the previous pretext tasks that focus on low-level image details, the contrastive learning encourages the model to learn useful semantic dynamics from previously learned motion features at a lower level, and is more favorable for motion learning at higher levels where the spatial and temporal resolutions of feature maps are low~\cite{oord2018representation,chen2020simple,han2019video}.
To acquire the preliminary motion cues to initialize the hierarchical motion learning, we exploit the video frame reconstruction~\cite{jason2016back} as an auxiliary task such that the whole motion representation learning enjoys a unified self-supervised setup.

We realize the proposed motion learning module via a side network branch, which is lightweight and flexible to be embedded into a variety of backbone CNNs.
As shown in Figure~\ref{fig:teaser}(b), the side branch (i.e., the shaded region) for self-supervised motion learning is discarded after training, and only the learned motion features are retained in the form of residual connections~\cite{he2016deep}.
Our hierarchical design also promotes the fusion of appearance and motion by integrating the motion features into the backbone network at multiple abstraction levels.
This multi-level fusion paradigm is unachievable for previous motion learning methods~\cite{diba2019dynamonet,zhu2018hidden} that depend solely on low-level motion supervisions.

We summarize our main contributions as follows. 
(1) We propose a new learning framework for motion representation learning from raw video frames.
(2) We advance contrastive learning to a hierarchical design that bridges the semantic gap between low-level motion cues and high-level recognition tasks.
(3) To our knowledge, this work provides the first attempt to empower contrastive learning in motion representation learning for large-scale video action recognition.
(4) Our approach achieves superior results on four benchmarks without relying on off-the-shelf motion features or supervised pre-training.

\section{Related Work}

\paragraph{Action Recognition and Motion Extraction.} 
A large family of video action research focuses on motion modeling~\cite{simonyan2014two,multilayer,budget-aware,res3d,xie2018rethinking,prernn}.
For example, the two-stream networks model temporal information by leveraging external motion inputs such as optical flow and frame differences~\cite{simonyan2014two,budget-aware}. 
3D CNNs~\cite{res3d,xie2018rethinking} and RNNs~\cite{multilayer,prernn} are also widely used to simultaneously model appearance and motion. 

To learn more explicit motion representations, ActionFlowNet~\cite{ng2018actionflownet} uses pre-computed optical flow as an additional supervision to encode motion together with appearance.
However, the approach requires the extraction of optical flow which is time-consuming and infeasible for large-scale datasets. 
Recently, TVNet~\cite{fan2018end} formulates the TV-L1 algorithm in a customized network layer,  producing optical flow like motion features to complement static appearance.
A differentiable representation flow layer is also developed in~\cite{piergiovanni2019representation}.
Inspired by the correlation layer in FlowNet~\cite{dosovitskiy2015flownet}, CorrNet~\cite{corrnet} and MSNet~\cite{kwon2020motionsqueeze} leverage the correlation operation to extract motion from convolutional features.
All these studies employ classification loss as an indirect supervision or mimic optical flow design to learn motion extraction, while our approach explicitly realizes motion learning through the proposed hierarchical contrastive learning in a fully self-supervised way.

\vspace{-3mm}
\paragraph{Self-Supervised Learning (SSL).} 
To take advantage of the abundant unlabeled videos, numerous methods have been developed for sequence learning by various self-supervisory signals, such as frame interpolation~\cite{niklaus2018context,diba2019dynamonet}, sequence ordering~\cite{lee2017unsupervised,fernando2017self}, speed prediction~\cite{benaim2020speednet,wang2020self}, future prediction~\cite{zhou2016view,lotter2016deep}, and cross-sensor motion regularization~\cite{pillar-motion}.
While these methods do not require pre-trained networks nor video annotations, 
they tend to focus on the low-level information and may not effectively capture the high-level temporal dynamics. 

Recently, contrastive learning has been explored for self-supervised learning of video representations~\cite{han2019video,han2020self,qian2020spatiotemporal,self2020Tao,yang2020video}. Different pretext tasks are designed to facilitate the learning of spatio-temporal representations, such as future prediction~\cite{han2019video,han2020memory}, augmentation invariance~\cite{yang2020video}, and multi-view co-training~\cite{han2020self}.
Unlike these existing works that leverage SSL to learn discriminative video representations, our work focuses on \textbf{extracting motion information} from raw video frames in a self-supervised manner. Our motion learning can be naturally integrated into the supervised training process and further boosts the supervised learning performance on large-scale datasets. This is not achievable for previous SSL methods because they only serve as a network pre-training process. 
\section{Approach}


Given the convolutional features $\{\mathcal{F}^0, ..., \mathcal{F}^{L-1} \}$ at different levels of a backbone network, our aim is to learn a \textbf{hierarchy} of motion representations $\left\{\mathcal{P}^0, ..., \mathcal{P}^{L-1} \right\}$ for the corresponding levels. Here, $L$ indicates the total number of abstraction levels.
As the first step, we employ video frame reconstruction as the supervision to obtain the preliminary motion cues $\mathcal{P}^0$, which function as a bootstrap for the following hierarchical motion learning.
With that, we progressively learn the motion features in a bottom-up manner.
At each level $l>0$, we learn the motion features $\mathcal{P}^l$ by enforcing them to predict the future motion features at the previous level $l-1$, as described in Sec.~\ref{sec:motion}.
We use the contrastive loss as an objective such that $\mathcal{P}^l$ is trained to capture semantic temporal dynamics from $\mathcal{P}^{l-1}$. 
The learned motion features at each level are integrated into a backbone network via residual connections to perform appearance and motion feature fusion: $\mathcal{Z}^l = \mathcal{F}^l + g^l(\mathcal{P}^l)$, where $g^l(\cdot)$ is used to match the feature dimensions.
After learning motion at all levels, we jointly train the whole network for action recognition in a multi-tasking manner, as presented in Sec.~\ref{sec:action}. 

\subsection{Self-Supervised Motion Learning}\label{sec:motion}

\begin{figure}[t]
\centering
\includegraphics[width=0.85\linewidth]{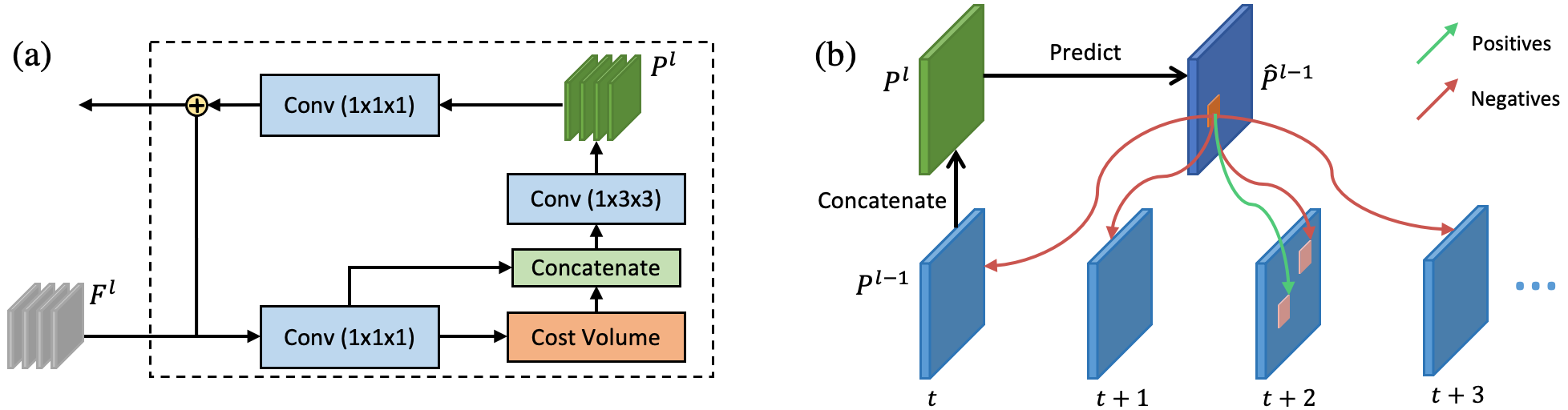}
\caption{(a) Architecture of the prime motion block. (b) Overview of our contrastive motion learning that adopts the higher-level motion features to predict the future ones at the lower level, as well as the illustration of positive and negative samples.\vspace{-2mm}}
\label{fig:method}\vspace{-2mm}
\end{figure}

\paragraph{Prime Motion Block.} 
We first introduce a lightweight prime motion block (PMB) to transform the convolutional features of a backbone network to more discriminative representations for motion learning. The key component of this block is a cost volume layer, which is inspired by the success of using cost volumes in stereo matching~\cite{hosni2012fast} and optical flow estimation~\cite{pwc-pami}.
An illustration of PMB is shown in Figure~\ref{fig:method} (a). 

Given a sequence of convolutional features $\mathcal{F} = \left \{F_0,...,F_{T-1} \right\}$ with length $T$, we first conduct a $1\!\times\! 1 \!\times\! 1$ convolution to reduce the input channels by $1/\beta$, denoted as $\tilde{\mathcal{F}}$.
This operation significantly reduces the computational overhead of prime motion block, and provides more compact representations to reserve the essential information to compute cost volumes.
The adjacent features are then re-organized to feature pairs $\mathcal{\tilde{F}}^* = \{ (\tilde{F}_0,\tilde{F}_1), ...,$
$(\tilde{F}_{T-2},\tilde{F}_{T-1}), (\tilde{F}_{T-1},\tilde{F}_{T-1}) \}$ for constructing the cost volumes.
The matching cost between two features is defined as:
$\text{cv}_t(x_1,y_1,x_2,y_2) = \text{sim}(\tilde{F}_t(x_1,y_1),  \tilde{F}_{t+1}(x_2,y_2))$,
where $\tilde{F}_t(x,y)$ denotes the feature vector at time $t$ and position $(x, y)$, and the cosine distance is used as the similarity function: $\text{sim}(u,v) = u^T v / \|u\| \|v\|$.
We replicate the last feature map $\tilde{F}_{T-1}$ to compute their cost volume in order to keep the original temporal resolution.
We limit the search range with the max displacement of $(x_2,y_2)$ to be $d$ and use a striding factor of $s$ to handle large displacements without increasing the computation.
As a result, the cost volume layer outputs a feature tensor of size $M\times H\times W$, where $M=(2\times\lfloor d/s\rfloor +1)^2$ and $H,W$ denote the height and width of a feature map.
Note that computing cost volumes is lightweight as it has no learnable parameters and much fewer FLOPs than 3D convolutions. 
Finally, we combine  cost volumes with  features obtained after dimension reduction, motivated by the observation that the two features are complementary  for localizing motion boundaries.

\paragraph{Preliminary Motion Cues.}
Although the prime motion block extracts rough motion features from convolutional features, we find that such features are easily biased towards appearance when jointly trained with the backbone network. 
Thus, an \textbf{explicit} motion supervision is vital for more effective motion learning at each level.

To initialize the progressive training, the preliminary motion cues, i.e., $\mathcal{P}^0$, are required as a bootstrap. 
They should encode some low-level but valid movement information to facilitate the following motion learning. 
Therefore, we adopt video frame reconstruction to guide the extraction of preliminary motion cues. 
This task can be formulated as a self-supervised optical flow estimation problem~\cite{jason2016back}, aiming to produce optical flow to allow frame reconstruction from neighboring frames.
Motivated by the success of recent work on estimating optical flow with CNNs~\cite{sun2018pwc}, we build a simple optical flow estimation module using 5 convolutional layers with dense connections.
We make use of the estimated optical flow to warp video frames through bilinear interpolation. The loss function consists of a photometric term that measures the error between the warped frame and the target frame, and a smoothness term that handles the aperture problem that causes ambiguity in motion estimation: $\mathcal{L}_\text{reconstruct} = \mathcal{L}_{\text{photometric}} + \zeta \mathcal{L}_{\text{smoothness}}$. 
We define the photometric error as:
\begin{equation}
\small
\label{eq:photo}
\hspace{-5pt}\mathcal{L}_{\text{photometric}} = \frac{1}{HWT} \sum_{t=1}^T\sum_{x=1}^W \sum_{y=1}^H  \mathbbm{1} \rho\left (I_t(x,y) - \hat{I}_t(x,y) \right ),\hspace{-1pt}
\end{equation}
where $\hat{I}_t$ indicates the warped frame at time $t$ and $\rho(z) = (z^2 + \epsilon^2)^{\alpha}$ is the generalized Charbonnier penalty function with $\alpha=0.45$ and $\epsilon=1e^{-3}$,
and the indicator function $\mathbbm{1}\in\left\{0,1\right\}$  excludes  invalid pixels that move out of the image boundary. 
Additionally, we compute the smoothness term as:
\begin{equation}
    \small
    \label{eq:smooth}
    \mathcal{L}_{\text{smoothness}} = \frac{1}{T} \sum_{t=1}^T\rho(\nabla_x U_t) + \rho(\nabla_y U_t) + \rho(\nabla_x V_t) + \rho(\nabla_y V_t),
\end{equation}
where $\nabla_x U/V$ and $\nabla_y U/V$ are the gradients of estimated flow fields $U/V$ in $x/y$ directions.

\paragraph{Hierarchical Motion Learning.}

Given above preliminary motion cues as a bootstrap, we propose to learn higher-level motion representations using a multi-level self-supervised objective based on the contrastive loss~\cite{chopra2005learning,hadsell2006dimensionality,gutmann2010noise}.
Our goal is to employ the higher-level motion features as a conditional input to guide the prediction of the future lower-level motion features that are well-learned from a previous step. By this way, the higher-level features are forced to understand a more abstract trajectory that summarizes motion dynamics from the lower-level ones.
This objective therefore allows us to extract slowly varying features that progressively correspond to high-level semantic concepts~\cite{oord2018representation,han2019video}.

Formally, let us denote the motion features generated by the prime motion block at level $l>0$ as $\mathcal{P}^l = \left\{P^l_0, ..., P^l_{T-1}\right\}$.
In order to train $\mathcal{P}^l$, we enforce $P^l_t$ to predict the future motion features at the previous level (i.e., $P^{l-1}_{>t}$), conditioned on the motion feature at the current time $P^{l-1}_t$, as illustrated in Figure~\ref{fig:method}(b). 
In practice, a predictive function $f_{\delta}$ is applied for the motion feature prediction at time step $t+\delta$: $\hat{P}^{l-1}_{t+\delta} = f_{\delta}([P^l_t, P^{l-1}_t])$, where $[\cdot, \cdot]$ denotes channel-wise concatenation.
We employ a multi-layer perception (MLP) with one hidden layer as the prediction function: $f_{\delta}(x) = W^{\text{(2)}}_{\delta} \sigma (W^{\text{(1)}}x)$, where $\sigma$ is ReLU and $W^{\text{(1)}}$ is shared across all prediction steps in order to leverage their common information. We define the objective function of each level as a contrastive loss, which encourages the predicted $\hat{P}^{l-1}$ to be close to the ground truth $P^{l-1}$ while being far away from the negative samples:
\begin{equation}
\small
\label{eq:cpc}
\hspace{-4pt}\mathcal{L}^l_{\text{contrastive}} = - \sum_{i \in \mathcal{S}} 
\left [ \text{log} \frac{\text{exp}(\text{sim}(\hat{P}^{l-1}_i, P^{l-1}_i) / \tau)} {\sum_{j \in \mathcal{S}} \text{exp}(\text{sim}(\hat{P}^{l-1}_i, P^{l-1}_j) / \tau)} \right ],
\end{equation}
where the similarity function is the same  cosine similarity used in computing cost volumes, and $\mathcal{S}$ denotes the sampling space of positive and negative samples. 

As shown in Figure~\ref{fig:method}(b), the positive sample of the predicted feature is the ground-truth feature that corresponds to the same video and locates at the same position in both space and time as the predicted one. 
Similar to~\cite{han2019video}, we define three types of negative samples for all prediction and ground-truth pairs: spatial negatives, temporal negatives and easy negatives. Considering efficiency, we randomly sample $N$ spatial locations for each video within a mini-batch to compute the loss.
Please see the supplementary material for more sampling details.
As illustrated in Figure~\ref{fig:teaser}(b), the contrastive motion learning is performed for multiple levels until the motion hierarchy of the whole network is built up.

\vspace{-3mm}
\paragraph{Progressive Training.}
Training the multi-level self-supervised learning framework simultaneously from the beginning is infeasible, as the lower-level motion features are initially not well-learned and the higher-level prediction would be arbitrary.
To facilitate the optimization process, we propose a progressive training strategy that learns motion features for one level at a time, propagating from low-level to high-level.
In practice, after the convergence of training at level $l-1$, we freeze all network parameters up to level $l-1$ (therefore fixing the motion features $\mathcal{P}^{l-1}$), and then start the training for level $l$.
In this way, the higher-level motion features can be stably trained with the well-learned lower-level ones.

\subsection{Joint Training for Action Recognition}
\label{sec:action}
Our ultimate goal is to improve video action recognition with the learned hierarchical motion features.
To integrate the learned motion features into a backbone network, we wrap our prime motion block into a residual block: $\mathcal{Z}^l = \mathcal{F}^l + g^l(\mathcal{P}^l)$, where $\mathcal{F}^l$ is the convolutional features at level $l$, $\mathcal{P}^l$ is the corresponding motion features obtained in Sec.~\ref{sec:motion}, 
and $g^l(\cdot)$ is a $1\!\times\!1\!\times\!1$ convolution. 
This seamless integration enables end-to-end fusion of appearance and motion information over multiple levels throughout a single unified network, instead of learning them disjointly like two-stream networks~\cite{simonyan2014two}. 
After the motion representations are self-supervised learned at all levels, we add in the classification loss to 
jointly optimize the total objective, which is a weighted sum of the following losses: 
\begin{equation}
\small
\label{eq:loss}
\mathcal{L}_{\text{total}} = \mathcal{L}_{\text{classification}} +  \lambda \mathcal{L}_{\text{reconstruct}} + \sum_l \gamma^l \mathcal{L}^l_{\text{contrastive}},
\end{equation}
where $\lambda$ and $\gamma^l$ are the weights to balance related loss terms. As illustrated in Figure~\ref{fig:teaser}(b), our multi-level self-supervised learning is performed through a side network branch, which can be flexibly embedded into standard CNNs. Furthermore, this self-supervised learning side branch is discarded after training so that our final network can well maintain the efficiency at runtime for inference.


\section{Experiments}



We evaluate the proposed approach on four benchmarks: Kinetics-400 (K400)~\cite{carreira2017quo}, Something-Something V1\&V2 (SS-V1\&V2)~\cite{goyal2017something} and UCF-101~\cite{soomro2012ucf101}.
Our motion learning module is generic and can be instantiated with various video networks~\cite{NonLocal2018,carreira2017quo,tran2018closer}. 
In our experiments, we use the standard networks R2D~\cite{NonLocal2018} and R(2+1)D~\cite{tran2018closer} as our backbones.
We follow the standard recipe in~\cite{feichtenhofer2019slowfast} for model training.
Note that all models are trained from scratch or self-supervised pre-trained without additional annotations or pre-computed optical flow. 
More details on datasets and implementations are available in the supplementary material.

\newcommand{\ta}[1]{\cellcolor[rgb]{0.95,0.95,.95}#1}
\newcommand{\tb}[1]{\cellcolor[rgb]{0.80,0.80,.80}#1}

\begin{figure}
\begin{floatrow}
\capbtabbox{
    \scriptsize
    \setlength{\tabcolsep}{4pt}
    \begin{tabular}{c|ccc|c}
    \hline
    \multirow{2}{*}{Input}  & \multicolumn{3}{c|}{Supervision} & \multirow{2}{*}{Score} \\
                        & Action & Reconstruction & Contrastive &  \\ \hline\hline
    \multirow{3}{*}{Level 1} & \ta{\cmark}           & \tb{}          & \ta{}            &   2.3    \\
                             & \ta{}            & \tb{\cmark}         & \ta{}           &   \textbf{3.1}    \\
                             & \ta{}            & \tb{}           & \ta{\cmark}    & -- \\ \hline
    \multirow{3}{*}{Level 2} & \ta{\cmark}    & \tb{}          & \ta{}            &  1.7     \\
                             & \ta{}             & \tb{\cmark}    & \ta{}           &   2.5    \\
                             & \ta{}            &  \tb{}         & \ta{\cmark}    &  \textbf{3.0}    \\ \hline
    \multirow{3}{*}{Level 3} & \ta{\cmark}     &    \tb{}       & \ta{}           &  2.4    \\
                             &  \ta{}           & \tb{\cmark}    & \ta{}           &  1.7     \\
                             &   \ta{}          &    \tb{}       &  \ta{\cmark}    & \textbf{3.0} \\ \hline
    \end{tabular}
    \setlength{\tabcolsep}{1.4pt}
}{
    \caption{Comparison of efficacy scores of the motion features learned at different levels under different supervisory forms.}
    \vspace{-2mm}
    \label{table:exp_score}
}
\ffigbox{
    \centering
    \includegraphics[height=3.5cm]{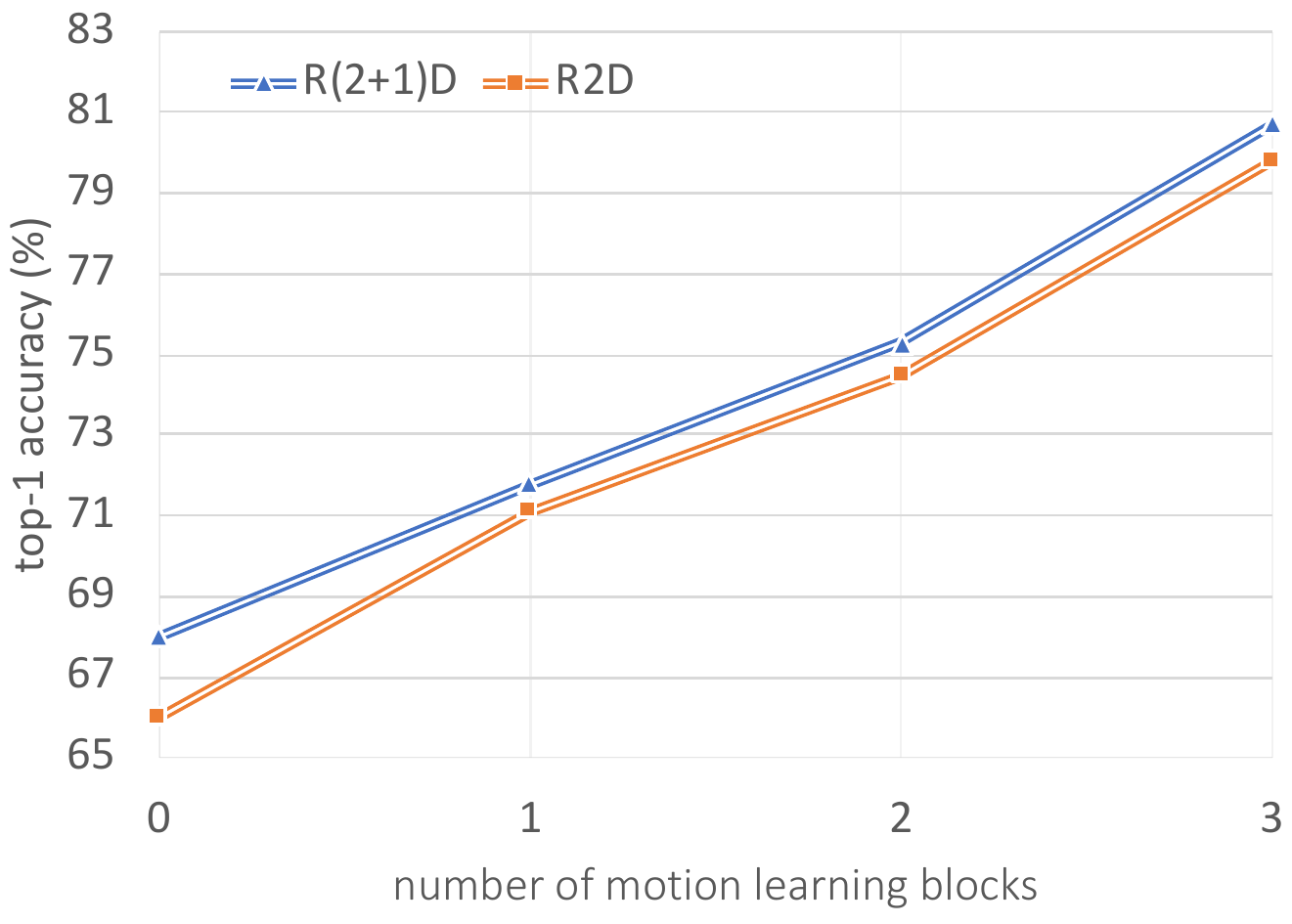}
}{
    \caption{Comparison of the top-1 accuracy on UCF-101 with incrementally adding the proposed motion learning blocks.} \vspace{-2mm}
    \label{fig:exp_self}
}
\end{floatrow}
\end{figure}

\subsection{Ablation Study}

\paragraph{Supervision for Motion Learning.}
\label{sec:exp_motion}
We first compare the motion features learned at different levels by different supervisions.
As our motion learning is based on the self-supervisions that are not directly related with the final action recognition, 
we first define a measurement to reflect the efficacy of the learned motion features.
Towards this goal, we take the extracted motion feature as input and train a lightweight classifier for action recognition on UCF-101.
We define the efficacy score as: $\text{Score} = \text{Acc}_{\text{train}} / (\text{Acc}_{\text{train}} - \text{Acc}_{\text{test}})$,
where $\text{Acc}_{\text{train}}$ and $\text{Acc}_{\text{test}}$ indicate the top-1 accuracy on the training and test sets.
Intuitively, a higher score implies that the representation is more discriminative (with higher training accuracy) and generalizes better (with a lower performance gap between training and testing).

\begin{figure}[t]
\centering
\includegraphics[width=0.85\linewidth]{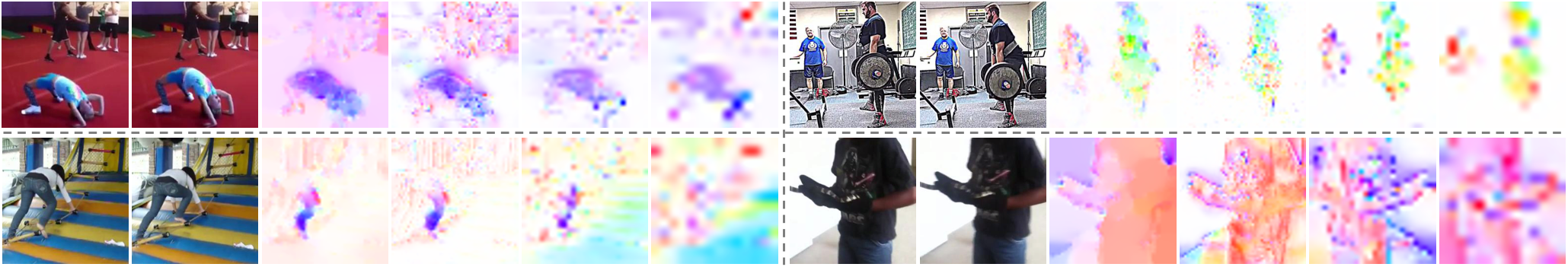}
\vspace{-2mm}
\caption{Visualization of the estimated optical flow at different feature abstraction levels. For each group, columns 1-2 are adjacent frames; column 3 is the reference optical flow extracted by~\cite{liu2009beyond}; columns 4-6 are the estimated optical flow at levels 1, 2 and 3. \vspace{-4mm}} 
\label{fig:flow}
\end{figure}

Table~\ref{table:exp_score} shows the efficacy scores of motion features at different levels with different supervisions, where
``action'' indicates the supervision by action classification, and ``reconstruction'' and ``contrastive'' refer to the supervisions by frame reconstruction and contrastive learning.
Levels 1, 2 and 3 correspond to the motion features extracted after $\texttt{res}_2$, $\texttt{res}_3$ and $\texttt{res}_4$ of R2D.
We observe that the self-supervision of low-level frame reconstruction is particularly effective at level 1, but its performance degrades dramatically at higher levels due to lower spatial/temporal resolutions and higher abstraction of convolutional features.
In contrast, the proposed self-supervision by hierarchical contrastive learning is more stable over different levels and more effective to model motion dynamics.
It is also observed that the self-supervision, with correct choices at different levels, consistently outperforms the supervision by action classification, which is consistent with the findings in~\cite{ng2018actionflownet,stroud2018d3d,diba2019dynamonet}.
In Figure~\ref{fig:flow}, we visualize the estimated optical flow, the by-product of frame reconstruction at each level, and find that more accurate optical flow indeed presents at lower levels.

\setlength{\tabcolsep}{6pt}
\begin{table}[t]
\scriptsize
\centering
\begin{tabular}{l|cc|c|ccc}
\toprule
\multicolumn{1}{c|}{Methods} & PMB & Self-Sup & FLOPs & UCF-101      & SS-V1    & K400     \\ \midrule
Baseline: R2D               &          &       & 1.00$\times$  & 66.0 / 86.0  & 36.1 / 68.1     & 64.8 / 85.1    \\
Ours: R2D                   &    \cmark       &    & 1.18$\times$     & 71.6 / 89.7   & 43.6 / 74.7    & 65.6 / 85.5  \\
Ours: R2D                   &       \cmark  &    \cmark   & 1.18$\times$    & \textbf{79.8} / \textbf{94.4} & \textbf{44.3} / \textbf{75.8} & \textbf{67.3} / \textbf{86.4} \\ \midrule
Baseline: R(2+1)D           &         &      & 1.00$\times$     & 68.0 / 88.2 & 48.5 / 78.1    & 66.8 / 86.6\\
Ours: R(2+1)D               &    \cmark       &     & 1.11$\times$    & 73.4 / 92.1  &  49.2 / 77.9    & 67.4 / 86.9 \\
Ours: R(2+1)D               &      \cmark   &    \cmark    &1.11$\times$    & \textbf{80.7} /  \textbf{95.6} &    \textbf{50.4} / \textbf{78.9}       & \textbf{68.3} / \textbf{87.4} \\ \bottomrule
\end{tabular}
\caption{Ablation study on the prime motion block (PMB) and self-supervision (Self-Sup) for action recognition. We report the computational cost and top-1 / top-5 accuracy (\%) on the three benchmarks. Models are evaluated using a single clip per video to eliminate the impact of test-time augmentation.}
\label{table:ablation}
\vspace{-2mm}
\end{table}
\setlength{\tabcolsep}{1.4pt}

\setlength{\tabcolsep}{6pt}
\begin{table}[t]
    \scriptsize
    \centering
    \begin{tabular}{l|cccc|ccc}
    \toprule
    \multicolumn{1}{c|}{Methods} & Pre-train & Flow & Arch. (Depth) & GFLOPs$\times$Views$^\dagger$ & K400  & SS-V1 & SS-V2 \\ \midrule
    I3D~\cite{carreira2017quo} & IN-1K & \cmark &  I3D & 216 $\times$ N/A & 75.7 &-- &--\\
    S3D-G~\cite{xie2018rethinking} & IN-1K & \cmark & S3D & 143 $\times$ N/A  &77.2 & 48.2 &-- \\        Non-local~\cite{NonLocal2018} & IN-1K & & NL R3D (50) & 282 $\times$ 30  & 76.5 & 44.4 & --\\
    R(2+1)D~\cite{tran2018closer} & Sports1M & \cmark & R(2+1)D (34) & 304 $\times$ 115 & 75.4 & -- &-- \\

    TSM~\cite{lin2019tsm} & IN-1K & & R2D (50) & 65 $\times$ 30 & 74.7 & 49.7 & 63.4 \\
    ECO~\cite{zolfaghari2018eco} & -- & \cmark & ECO$_{\text{En}}$ & N/A &70.0 & 49.5 &-- \\
    SlowFast~\cite{feichtenhofer2019slowfast} & -- &  & SlowFast (101) & 106 $\times$ 30 & 77.9 &-- &--\\ \midrule
    Disentangling~\cite{zhao2018recognize} & IN-1K & & Disentangling & N/A &71.5 & --& --\\
    D3D~\cite{stroud2018d3d} & IN-1K & & S3D & N/A & 75.9 &-- &-- \\
    STM~\cite{stm} & IN-1K & & R2D (50) & 66.5 $\times$ 30 & 73.7 & 50.7 & 64.2 \\
    Rep. Flow~\cite{piergiovanni2019representation} & -- & & R(2+1)D (50) & N/A & 77.1 &-- &-- \\
    MARS~\cite{crasto2019mars} & -- & & 3D ResNeXt (101) & N/A& 72.7 &--&-- \\
    DynamoNet~\cite{diba2019dynamonet} & -- & & STCnet (101) & N/A & 77.9 &--&-- \\ \midrule \midrule
    \textbf{Ours} & -- & &R2D (50) & 49 $\times$ 30  &74.8 & 46.2 & 59.4 \\
    \textbf{Ours} & -- & & R(2+1)D (101) & 150 $\times$ 30 & \textbf{78.3} & \textbf{52.8} & \textbf{64.4} \\ \bottomrule
    \end{tabular}
    \vspace{-2mm}
    \caption{Comparison of the top-1 accuracy (\%) with the state-of-the-art methods on Kinetics-400 (K400) and Something-Something V1\&V2 (SS-V1\&V2). IN-1K indicates the ImageNet-1K dataset. $^\dagger$ GFLOPs and views are reported for the experiments on K400. } 
    \label{table:sota_kinetics}
\end{table}

\paragraph{Contributions of Individual Components.}
We verify the contributions of the proposed components in Table~\ref{table:ablation}. 
It is obvious that our approach consistently and significantly improves the action recognition accuracy for both 2D and 3D action networks.
Our prime motion block provides complementary motion features at multiple levels, and the self-supervision further enhances the representations to encode semantic dynamics.
In particular, for the dataset that heavily depends on temporal information like SS-V1, our approach remarkably improves the performance of baseline R2D by 8.2\%. 
For the dataset that is small-scale and tends to overfit to the appearance information like UCF-101, our method improves model generalization and achieves 13.8\% improvement. Moreover, our motion learning module only introduces a small overhead to FLOPs of the backbone network.   

We next validate the contribution of our motion learning at each level by incrementally adding the proposed motion feature learning block to the baseline.
Figure~\ref{fig:exp_self} demonstrates the results based on the backbones of R2D and R(2+1)D on UCF-101.
We observe that notable gains can be obtained at multiple levels, and the performance gain does not vanish with the increase of motion learning blocks, suggesting the importance of leveraging hierarchical motion information across all levels.

\begin{table}
\begin{floatrow}
\capbtabbox{
    \scriptsize
    \centering
    \begin{tabular}{l|c|c}
    \toprule
    \multicolumn{1}{c|}{Methods} & Dataset & Acc.  \\ \midrule
    TSN~\cite{wang2016temporal} & IN-1K + K400 & 91.7 \\
    TSM~\cite{lin2019tsm} & IN-1K + K400 & 95.9 \\
    I3D~\cite{carreira2017quo} & IN-1K + K400 & 95.4 \\
    S3D-G~\cite{xie2018rethinking} & IN-1K + K400 & 96.8  \\
    LGD-3D~\cite{qiu2019learning} & IN-1K + K600 & 97.0 \\ 
    Disentangling~\cite{zhao2018recognize} & IN-1K + K400 & 95.9 \\
    D3D~\cite{stroud2018d3d} & IN-1K + K400 &  97.0 \\
    STM~\cite{stm} & IN-1K + K400 & 96.2 \\
    DynamoNet~\cite{diba2019dynamonet} & YT-8M + K400 & \textbf{97.8} \\ 
    R(2+1)D~\cite{tran2018closer} & K400 & 96.8  \\
    MARS~\cite{crasto2019mars} & K400 & 97.0 \\
    \midrule \midrule
    \textbf{Ours}, R(2+1)D-101 & K400 & \textbf{97.8} \\ \bottomrule
    \end{tabular}
    \vspace{-2mm}
}{
\caption{Comparison with \textbf{supervised} learning methods on UCF-101 (3 splits). Models are first supervisedly pre-trained on large-scale datasets and then fine-tuned on UCF-101. \vspace{-2mm}}
\label{table:sota_ucf}
}
\capbtabbox{
    \setlength{\tabcolsep}{3pt}
    \scriptsize
    \begin{tabular}{l|c c c|c}
    \toprule
    \multicolumn{1}{c|}{Method} & Dataset  & Res. & Arch. (depth)                      & Acc. \\ \midrule
    OPN~\cite{lee2017unsupervised}        & UCF  & 224  & VGG (14)   &      59.8     \\
    ActionFlowNet$\dagger$~\cite{ng2018actionflownet}       & UCF  & 224 & R3D (18)    & 83.9     \\
    DynamoNet~\cite{diba2019dynamonet}        & YT-8M & 112 & STCNet (133)    &     88.1     \\ 
    3D-Puzzle~\cite{kim2019self}      & K400  & 224 & R3D (17)    & 63.9     \\
    DPC~\cite{han2019video}     & K400  & 128 & R-2D3D (33)    &  75.7     \\
    SpeedNet~\cite{benaim2020speednet} & K400  & 224 & S3D-G (23) & 81.1 \\
    PacePred~\cite{wang2020self} & K400 &  112 & R(2+1)D (18) & 77.1 \\
    MemDPC-RGB~\cite{han2020memory} & K400 &  224  & R-2D3D (33) & 78.1 \\
    CVRL~\cite{qian2020spatiotemporal} & K400  & 224 & R3D (50) & 92.1 \\
    CoCLR-RGB$\dagger$~\cite{han2020self} & K400 &  128 & S3D (23) & 87.9 \\
    \midrule \midrule
    \textbf{Ours} (random init.)     & --       & 128 & S3D (23)  & 77.0 \\
    \textbf{Ours}     & K400  &    128 &  S3D (23)  &    87.1     \\ \bottomrule
    
    \end{tabular}
    \vspace{-2mm}
}{
    \caption{Comparison with the \textbf{self-supervised} methods on UCF-101 (split-1). Duration of the pre-training datasets: UCF (1d), K400 (28d), YT-8M (1y). $\dagger$ denotes the methods requiring pre-extracted optical flow for model training. \vspace{-2mm}}
    \label{table:sota_self}
}
\end{floatrow}
\end{table}

\subsection{Comparison with State-of-the-Art}
\label{sec:exp_sota}
\paragraph{Supervised Action Recognition.} We compare our approach with the state-of-the-art methods on the four action recognition benchmarks.
Table~\ref{table:sota_kinetics} shows the comparisons on K400 and SS-V1\&V2.
Without using optical flow or supervised pre-training, our model based on backbone R(2+1)D-101 achieves the best results among the single-stream methods over all three datasets.
Our approach also outperforms most two-stream methods, apart from the recent two-stream TSM~\cite{lin2019tsm} on SS-V2.
As for the datasets that focus more on temporal modeling like SS-V1\&V2, 2D networks are usually not able to achieve as good results as 3D models.
However, by equipping with the proposed motion learning module, we find that our method based on backbone R2D-50 outperforms some 3D models, such as R(2+1)D and NL I3D.
Our approach also achieves superior results compared with the most recent work that are specifically designed for temporal motion modeling (i.e., the second group in Table~\ref{table:sota_kinetics}).
More importantly, our motion learning is fully self-supervised from raw video frames without any supervisions from optical flow or pre-trained temporal stream. 




We also conduct experiments on UCF-101 to demonstrate the transferability of our learned features to small-scale datasets. Following the standard setting in previous work~\cite{carreira2017quo,qiu2019learning}, we pre-trained our models on K400 for action classification and then finetune the weights on UCF-101.
We report the average accuracy over all 3 splits. As shown in Table~\ref{table:sota_ucf}, our approach achieves 97.8\% top-1 accuracy, comparable with the state-of-the-art results among the single-stream methods on UCF-101.

\vspace{-2mm}
\paragraph{Self-Supervised Pre-Training.} Self-supervised learning of video representations has been gaining increasing attention in recent years~\cite{fernando2017self,lee2017unsupervised,kim2019self,han2019video,diba2019dynamonet}.
In addition to hierarchical motion learning, our approach can also serve as pre-training of a network.
As an example, after self-supervised motion learning on K400 (without using its action labels), we fine-tune the network on UCF-101 for action recognition, as shown in Table~\ref{table:sota_self}.
For fair comparisons, we adopt the S3D network as backbone and follow the same experimental setting as used in CoCLR~\cite{han2019video}. 
Interestingly, our approach is capable of learning effective video representation that is comparable with state-of-the-art self-supervised learning methods, even though network pre-training is \textbf{not} the main focus of our work.
Note that previous work requires pre-extracted optical flow for model training~\cite{ng2018actionflownet,han2020self} or much larger pre-training datasets such as YouTube8M (YT-8M)~\cite{diba2019dynamonet} to achieve state-of-the-art results.

\begin{figure}
\begin{floatrow}
\ffigbox{
    \centering
    \includegraphics[width=0.78\linewidth]{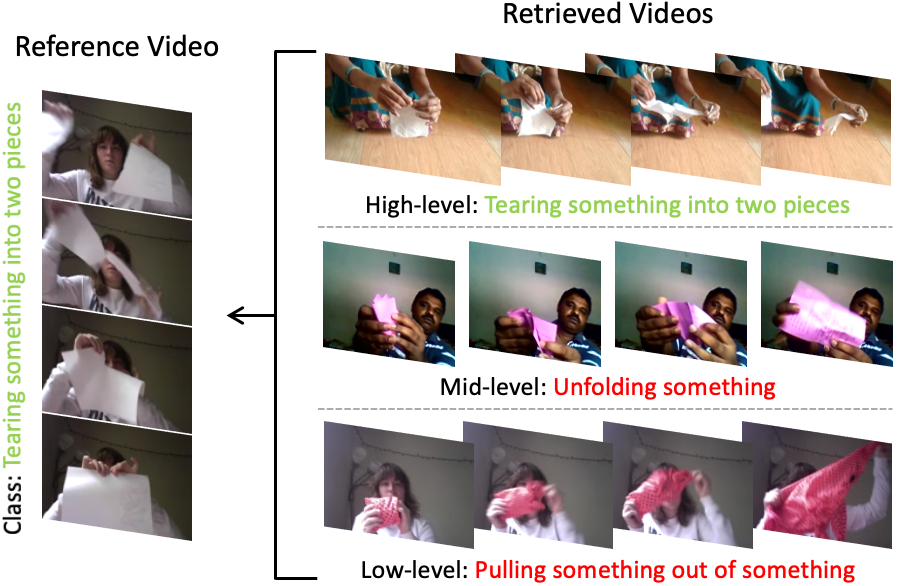}
    \vspace{-2mm}
}{
    \caption{Examples of the retrieved videos to reflect the different motion semantics learned by the motion features at different levels. \vspace{-2mm}}
    \label{fig:retrieval}
}
\capbtabbox{
    \centering
    \scriptsize
    \renewcommand{\arraystretch}{1.2}
    \setlength{\tabcolsep}{4pt} 
    \begin{tabular}{l | l}
        \toprule
        Motion  & Top-2 Classes with Largest Relative Gains \\
        \midrule
        Low-  & ``turning the camera left" ($73.8$)\\
        level  & ``turning the camera right" ($71.7$)\\
        \midrule
        Mid-  & ``showing smth to the camera" ($0\rightarrow19.4$)\\
        level  & ``showing smth behind smth" ($0\rightarrow12.2$)\\
        \midrule
        High- & ``pulling two ends of smth" ($0\rightarrow14.0$)\\
        level  & ``tipping smth with smth in it" ($0\rightarrow12.5$)\\
        \bottomrule
    \end{tabular}\vspace{-2mm}
}{
    \caption{Comparison of video classification accuracy using the motion features learned at different levels. \vspace{-2mm}}
    \label{tab:knn}
}
\end{floatrow}
\end{figure}

\subsection{Analysis on Motion Features at Different Levels}
Here we investigate the different semantics integrated in the motion features learned at different levels.
We first conduct a video retrieval experiment using the learned motion features and find that motion features at lower levels tend to retrieve the videos sharing similar elementary movements but lacking of high-level action correlation, as illustrated in Figure~\ref{fig:retrieval}.

Based on the retrieval results, we can further build a k-nearest-neighbor (KNN) classifier that assigns the query video to the majority action class among its top k nearest neighbors.  
As shown in Table~\ref{tab:knn}, using low-level motion features can obtain relatively high accuracy for some action classes with apparent moving patterns (e.g., ``turning the camera left"). This indicates that the low-level motion features are capable of extracting elementary movements from raw video frames. On the other hand, motion features at higher levels can recognize actions that require finer understanding of high-level motion semantics (e.g., ``pulling two ends of smth").
More details are available in the supplementary material.

\section{Conclusion}

We have presented hierarchical contrastive motion learning, a multi-level self-supervised framework that progressively learns a hierarchy of motion features from raw video frames.
A discriminative contrastive loss at each level provides explicit self-supervision for motion learning.
This hierarchical design bridges the semantic gap between low-level motion cues and high-level recognition tasks, meanwhile promotes effective fusion of appearance and motion information to finally boost action recognition.
%
%
%
%
Extensive experiments on four benchmarks show that our approach compares favorably against the state-of-the-art methods yet without requiring optical flow or supervised pre-training.

\appendix
\section*{Appendix}
Section~\ref{sec:2} shows qualitative results using Grad-CAM++. Sections~\ref{sec:3} and \ref{sec:4} respectively provide more details of our approach as well as the training on each specific dataset.

\section{Qualitative Result}
\label{sec:2}
\begin{figure}[t]
\centering
\includegraphics[width=\linewidth]{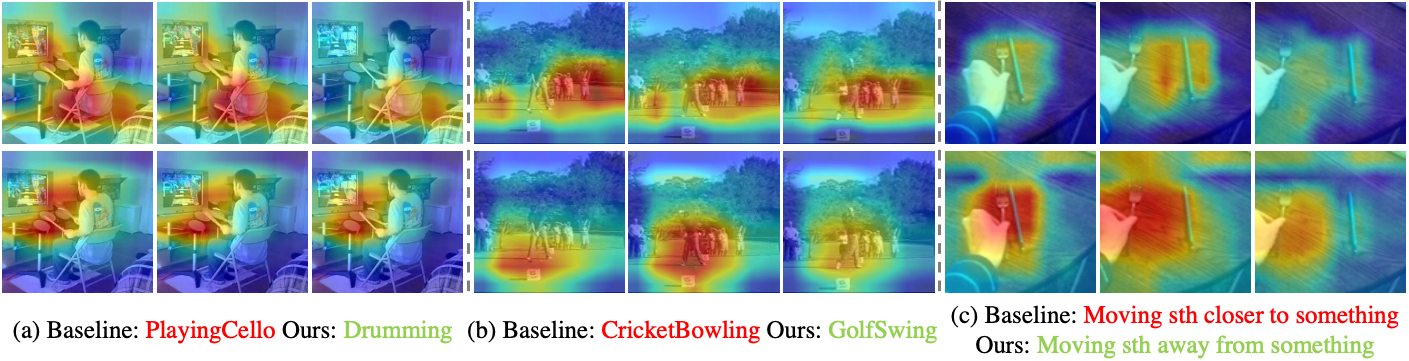}
\caption{Visualization of the learned features by Grad-CAM++~\cite{chattopadhyay2017grad}. Fonts in green and red indicate correct recognition and misclassification. (a-b) Features learned by our approach (bottom) are more sensitive to the regions with important motion cues. (c) Our motion learning module (bottom) equips the 2D backbone with the ability of reasoning the temporal order of video frames.}
\label{fig:visualization}
\end{figure}

To qualitatively verify the impact of the learned motion features, we utilize Grad-CAM++~\cite{chattopadhyay2017grad} to visualize the class activation map of the last convolution layer.
Figure~\ref{fig:visualization} shows the comparison between baseline and our model with the backbone R2D-50 on UCF-101 and Something-V1.
Our model attentions more on the regions with informative motion, while the baseline tends to be distracted by the static appearance.
For instance, in Figure~\ref{fig:visualization}(a), our method focuses on the moving hands of the person, while the baseline concentrates on the static human body.
Our motion learning module also equips the 2D backbone with effective temporal modeling ability.
As shown in Figure~\ref{fig:visualization}(c), our model is capable of reasoning the temporal order of the video and predicting the correct action, while the baseline outputs the opposite prediction result as it fails to capture the chronological relationship.

\section{Model Details}
\label{sec:3}
\subsection{Backbones}
\label{sec:backbone}

We adopt the standard convolutional networks R2D-50~\cite{NonLocal2018} and R(2+1)D-101~\cite{tran2018closer} as the backbones used in our experiments. A few changes are made to improve the computation efficiency, as demonstrated in Table~\ref{table:backbone}. Compared with the original network R(2+1)D-101, our backbone supports higher input resolution and applies bottleneck layers with consistent number of channels. We start temporal striding from $\texttt{res}_3$ rather than $\texttt{res}_4$, and employ the top-heavy design as used in~\cite{xie2018rethinking} for R(2+1)D-101, i.e., only using temporal convolutions in $\texttt{res}_4$ and $\texttt{res}_5$.

\newcommand{\resblockS}[3]{\multirow{4}{*}{\(\left[\begin{array}{c}\text{1$\times$1$\times$1, #1}\\[-.1em] \text{1$\times$3$\times$3, #1}\\[-.1em] \text{1$\times$1$\times$1, #2} \end{array}\right]\)$\times$#3}
}
\newcommand{\resblockT}[4]{\multirow{4}{*}{\(\left[\begin{array}{c}\text{1$\times$1$\times$1, #1}\\[-.1em] \text{(3$\times$1$\times$1, #1)}\\[-.1em] \text{1$\times$3$\times$3, #1}\\[-.1em] \text{1$\times$1$\times$1, #2} \end{array}\right]\)$\times$#3 (#4)}
}
\newcommand{\resblockTT}[3]{\multirow{4}{*}{\(\left[\begin{array}{c}\text{1$\times$1$\times$1, #1}\\[-.1em] \text{(3$\times$1$\times$1, #1)}\\[-.1em] \text{1$\times$3$\times$3, #1}\\[-.1em] \text{1$\times$1$\times$1, #2} \end{array}\right]\)$\times$#3}
}
\newcommand{\convblock}[1]{\multirow{2}{*}{\(\begin{array}{c}\text{1$\times$7$\times$7, #1}\\[-.1em] \text{stride: 1$\times$2$\times$2} \end{array}\)}
}
\newlength\savewidth\newcommand\shline{\noalign{\global\savewidth\arrayrulewidth\global\arrayrulewidth 1pt}\hline\noalign{\global\arrayrulewidth\savewidth}}

\begin{table}
    \scriptsize
    \centering
    \setlength{\tabcolsep}{6pt}
    \begin{tabular}{c l l}
    Layer & \multicolumn{1}{c}{R2D-50 / R(2+1)D-101} & Output Size \\
    \shline 
    \texttt{input} & \multicolumn{1}{c}{--}  & T $\times$ 224 $\times$ 224 \\
    \hline
    \multirow{2}{*}{$\texttt{conv}_{\text{1}}$} & \convblock{64} & \multirow{2}{*}{T $\times$ 112 $\times$ 112} \\
      &  &  \\
    \hline
    \multirow{4}{*}{$\texttt{res}_{\text{2}}$} & \resblockS{64}{256}{3}  & \multirow{4}{*}{ T $\times$ 56 $\times$ 56}\\
      &  &  \\
      &  &  \\
      &  &  \\
    \hline
    \multirow{4}{*}{$\texttt{res}_{\text{3}}$} & \resblockS{128}{512}{4} & \multirow{4}{*}{ $\text{T}/2$ $\times$ 28 $\times$ 28}\\
      &  &  \\
      &  &  \\
      &  &  \\
    \hline
    \multirow{4}{*}{$\texttt{res}_{\text{4}}$} & \resblockTT{256}{1024}{6 / 23}  & \multirow{4}{*}{ $\text{T}/4$ $\times$ 14 $\times$ 14}\\
      &  &  \\
      &  &  \\
      &  &  \\
    \hline
    \multirow{4}{*}{$\texttt{res}_{\text{5}}$} & \resblockTT{512}{2048}{3}  & \multirow{4}{*}{ $\text{T}/4$ $\times$ 7 $\times$ 7}\\
      &  &  \\
      &  &  \\
      &  &  \\ 
    \shline
    \end{tabular}
    \caption{Details of the architectures of the backbone networks R2D-50 / R(2+1)D-101 used in our experiments.}
    \label{table:backbone}
\end{table}

\subsection{Sampling Strategy}
We denote the predicted motion feature at level $l$ as $\hat{P}^l_{t,k}$, where  $t\in\left\{1, ...,T^l\right\}$ is the temporal index, and $k\in\left\{(1,1),(1,2),...(H^l,W^l)\right\}$ is the spatial index.
The only positive pair is $(\hat{P}^l_{t,k}, P^l_{t,k})$, which is the ground-truth feature that corresponds to the same video and locates at the same position in both space and time as the predicted one.
Following the terminology used in~\cite{han2019video}, we define three types of negative samples for all the prediction and ground-truth pairs $(\hat{P}^l_{t,k}, P^l_{\tau,m})$:

\noindent\textbf{Spatial negatives} are the ground-truth features that come from the same video of the predicted one but at a different spatial position, i.e., $k\neq m$. Considering the efficiency, we randomly sample $N$ spatial locations for each video within a
mini-batch to compute the loss. So the number of spatial negatives is $(N-1)T^l$.

\noindent\textbf{Temporal negatives} are the ground-truth features that come from the same video and same spatial position, but from different time steps, i.e., $k=m, t\neq \tau$. They are the hardest negative samples to classify, and the number of temporal negatives are $T^l-1$.

\noindent\textbf{Easy negatives} are the ground-truth features that come from different videos, and the number of easy negatives are $(B-1)NT^l$, where $B$ is the batch size.

\section{Experimental Details}
\label{sec:4}

\subsection{Datasets}
We extensively evaluate our proposed approach on the four benchmarks: Kinetics-400~\cite{carreira2017quo}, Something-Something (V1\&V2)~\cite{goyal2017something} and UCF-101~\cite{soomro2012ucf101}.
\textbf{Kinetics-400} is a large-scale video dataset with 400 action categories.
As some videos are deleted by their owners over time, our experiments are conducted on a subset of the original dataset with approximately 238K training videos ($\sim$96\%) and 196K validation videos ($\sim$98\%).
In practice, we notice a bit accuracy drop for the same model using our collected dataset with fewer training samples, suggesting that our results can be further improved with the full original dataset. 
\textbf{Something-Something (V1\&V2)} are 
more sensitive to temporal motion modeling.
Something-V1 contains about 100K videos covering 174 classes, and Something-V2 increases videos to 221K and improves video resolution and annotation quality.
\textbf{UCF-101} includes about 13K videos with 101 classes.
As the number of training videos is small, it is often used for evaluating unsupervised representation learning~\cite{diba2019dynamonet,ng2018actionflownet} and transfer learning~\cite{qiu2019learning,stroud2018d3d}.

\subsection{Implementation details}

We use the synchronized SGD with a cosine learning rate scheduling~\cite{loshchilov2016sgdr} and a linear warm-up~\cite{goyal2017accurate} for all model training.
The spatial size of the input is $224\times 224$, randomly cropped and horizontally flipped from a scaled video with the shorter side randomly sampled in [256, 320] pixels for R2D-50, and [256, 340] pixels for R(2+1)D-101.
We apply temporal jittering when sampling clips from a video.
The balancing weights for the joint training in Eq. (5) are set to $\lambda=15, \gamma^{1}=\gamma^{2}=0.25$, respectively.
We describe the training details for different benchmarks as follow.

\noindent\textbf{Kinetics-400.}
We sample a clip of $T = 16$ frames with a temporal stride 2 for the experiments using the backbone R2D-50 and $T = 32$ frames for those with the backbone R(2+1)D-101.
We train all models using the distributed SGD on GPU clusters with 8 clips per GPU. 
We set the learning rate per GPU to 0.0025, and linearly scale the learning rate according to the number of GPUs.
For the self-supervised training phase, all models are trained for 80 epochs with the first 10 epochs for warm-up, and the global batch normalization (BN)~\cite{ioffe2015batch} is used to avoid trivial solution.
As for the joint training phase, the models are trained for 200 epochs with the first 40 epochs for warm-up, and BN statistics is computed within each 8 clips. 

\noindent\textbf{Something-V1\&V2.}
Since this dataset has a higher frame rate than Kinetics-400, we sample a clip of $T = 32$ frames with a temporal stride 1 for all experiments. Models are trained for 150 epochs with the first 50 epochs for warm-up and the learning rate per GPU is also 0.0025.

\noindent\textbf{UCF-101.}
For the experiments described in Table~\ref{table:sota_ucf} in the supplementary material, the models are initialized with the weights pre-trained on Kinetics-400 for classification, and then are fine-tuned for 30 epochs with a batch size of 32 and a learning rate of 0.002.

\bibliography{egbib}
\end{document}